\documentclass[runningheads]{llncs}

\usepackage[T1]{fontenc}
\usepackage[utf8]{inputenc}
\usepackage{lmodern}

\usepackage{graphicx}
\usepackage{wrapfig}
\graphicspath{{fig/}{../fig/}}
\usepackage{booktabs}
\usepackage{multirow}
\usepackage{float}
\usepackage{xcolor}
\usepackage{pifont}
\usepackage{makecell}
\usepackage{wrapfig}
\usepackage{amsmath,amssymb}
\usepackage{hyperref}
\usepackage{tikz}
\usepackage{url}
\usetikzlibrary{arrows.meta, positioning, fit, backgrounds, calc}

\usepackage{hyperref}
\usepackage{cleveref}
\usepackage{enumitem}
\usepackage[misc]{ifsym}
\newcommand{\corr}{(\Letter)}

\title{The Reliability Gap in Benchmark Auditing: Distribution Shift and Scale as Failure Modes of Contamination Detection}
\titlerunning{Reliability Gap in Benchmark Auditing}
\toctitle{The Reliability Gap in Benchmark Auditing: Distribution Shift and Scale as Failure Modes of Contamination Detection}
\tocauthor{Wojciech Zarzecki, Jan Dubiński, Sebastian Cygert}

\author{Wojciech Zarzecki \inst{1,2}\thanks{This work was conducted mostly while the author was at NASK National Research Institute.} \corr \and
Jan Dubiński\inst{1,2} \and
Sebastian Cygert \inst{1,3}}
\authorrunning{W. Zarzecki et al.}

\institute{NASK National Research Institute, Warsaw, Poland \and
Warsaw University of Technology, Warsaw, Poland  \\
\email{wojciechzarzecki5@gmail.com} \and
Gdańsk University of Technology, Gdańsk, Poland}

\definecolor{matchgreen}{RGB}{34, 139, 34}
\definecolor{matchred}{RGB}{205, 92, 92}
\definecolor{matchorange}{RGB}{255, 140, 0}

\definecolor{mblue}{RGB}{66, 133, 244}
\definecolor{morange}{RGB}{245, 166, 35}
\definecolor{mgreen}{RGB}{76, 175, 80}
\definecolor{mpurple}{RGB}{149, 117, 205}

\newcommand{\bplus}[1]{\textcolor{#1}{\scalebox{1.2}{$\mathbf{+}$}}}
\newcommand{\bminus}[1]{\textcolor{#1}{\scalebox{1.2}{$\mathbf{-}$}}}
\newcommand{\bquest}[1]{\textcolor{#1}{\scalebox{1.2}{$\mathbf{?}$}}}

\newcommand{\gplus}{\bplus{matchgreen}}
\newcommand{\gminus}{\bminus{matchgreen}}
\newcommand{\rplus}{\bplus{matchred}}
\newcommand{\rminus}{\bminus{matchred}}
\newcommand{\oquest}{\bquest{matchorange}}

\newcommand{\redcross}[1][]{\textcolor{red}{\ding{55}}}
\newcommand{\greentick}[1][]{\textcolor{green}{\ding{51}}}

\begin{document}

\maketitle

\begin{abstract}

Benchmark contamination, where evaluation examples appear in a model's training data, threatens the validity of LLM assessment. Statistical tools for detecting training-data membership exist, but have been validated almost exclusively in controlled academic regimes: large, homogeneous pre-training corpora and transparent, single-stage training pipelines. 
Whether these methods remain reliable in realistic auditing scenarios remains unclear. We identify two under-studied failure modes: distribution shift, which arises when suspect and validation sets violate the IID assumption, and scale constraints, which arise because benchmarks are orders of magnitude smaller than pre-training corpora. We systematically evaluate three leading paradigms: LLM Dataset Inference, Post-Hoc Dataset Inference, and CoDeC across 25 models from multiple families (including Pythia, OLMo~2, and specialised cultural and medical LLMs) and scales (up to 27B). We then further extend our analysis to frontier industry models. Across 335 evaluations, only 201 yield correct outcomes. LLM Dataset Inference results in false positives under distribution shift, Post-Hoc Dataset Inference is underpowered at benchmark scale, and CoDeC provides only coarse provenance signals that are insufficient to verify individual benchmark splits. Our results reveal a systematic reliability gap between controlled validation and practical benchmark auditing, and show that statistical detection cannot yet replace transparent data provenance. We \href{https://anonymous.4open.science/r/reliability-gap-benchmark-auditing/README.md}{open-source} our benchmark for further research.

\end{abstract}

\section{Introduction}

    \begin{figure}
    \centering
    \includegraphics[width=1\linewidth]{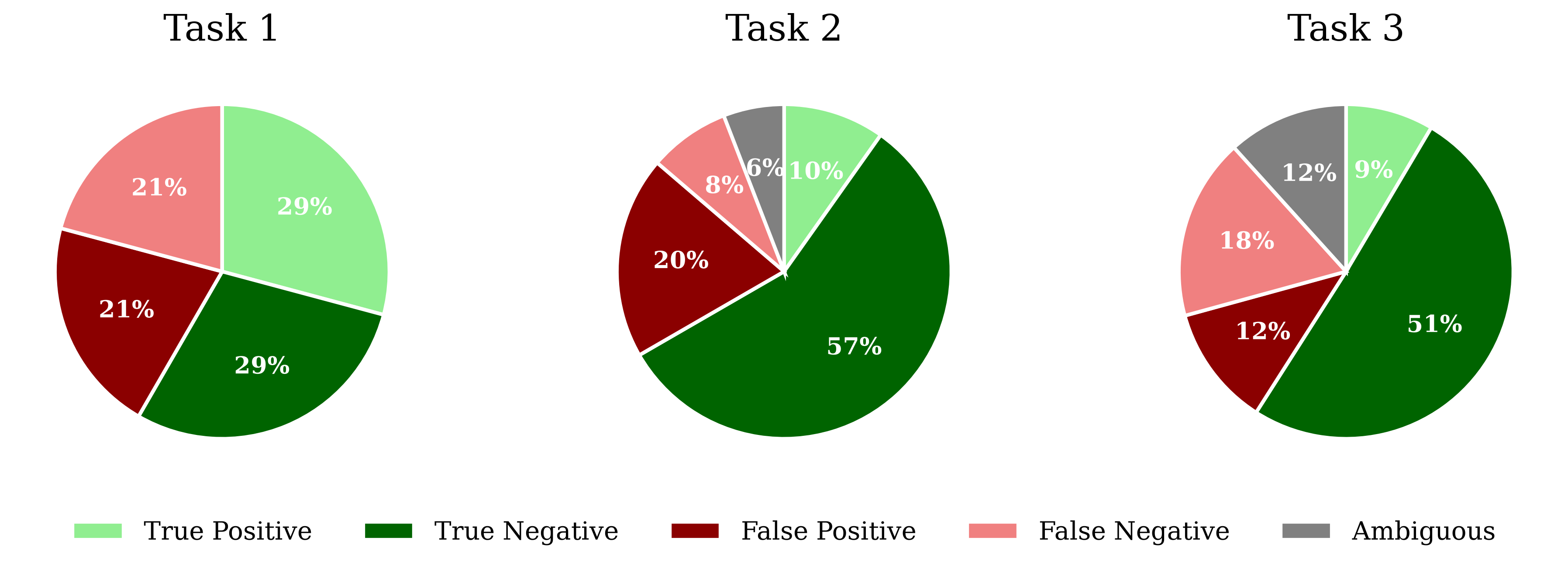}
     \caption{Summary of our evaluation of methods for detecting whether a model was trained on a benchmark, across three tasks:
    Task 1 evaluates vulnerability to limited reference data (\Cref{tab:task1_results});
    Task 2 evaluates split-level benchmark exposure on instruction-tuned OLMo 2 (\Cref{tab:task2_results});
    and Task 3 evaluates specialised post-training datasets, for medical QA and Polish-language LLMs (\Cref{tab:medical-contamination,tab:pllum-sft-contamination}).
    The counts aggregate detection outcomes from these tables.
    Overall, only 201 out of 335 (60\%) of the detection challenges are solved, suggesting that current approaches are not yet reliable for certifying benchmark integrity.
    }
    \label{fig:pie_chart}
    \end{figure}

In the era of rapidly scaling large language models, both model size and training data volume have increased by orders of magnitude. This scaling has unlocked new capabilities in reasoning, mathematics, and multi-domain knowledge, pushing performance on many public benchmarks to saturation.
At the same time, benchmark contamination has emerged as a serious threat to the validity of LLM evaluation. 
 
When benchmark splits leak into scraped or aggregated corpora, benchmark performance may reflect prior exposure rather than genuine generalization. As models saturate widely used benchmarks, distinguishing between reasoning capability and memorization has become a central challenge of modern LLM assessment.

Existing literature addresses this issue by developing principled statistical tools for detecting training data, including \textit{LLM Dataset Inference}~\cite{maini2024llmdatasetinferencedid}, \textit{Post-Hoc Dataset Inference}~\cite{zhao2025posthocdi}, and direct contamination measures such as \textit{CoDeC}~\cite{zawalski2025detectingdatacontaminationllms}. However, these methods are primarily validated in restricted academic settings, for example using the Pythia~\cite{biderman2023pythiasuiteanalyzinglarge} suite trained on The Pile~\cite{gao2020pile800gbdatasetdiverse}, where full data transparency and strict distributional assumptions hold. Academic model suites typically rely on relatively simple, single-stage pre-training on large, homogeneous corpora, which simplifies contamination analysis.

It remains unclear whether these methods remain reliable in more realistic settings. In practice, benchmark auditing concerns evaluation sets that are orders of magnitude smaller than pre-training corpora, and models that undergo multiple post-training stages, including instruction tuning on curated mixtures. To study this beyond academic model suites, our evaluation includes 21 openly available instruction-tuned models with sizes reaching 27B parameters (plus 4 pre-trained-only models, for 25 in total).
Therefore, in this work, we investigate whether existing detection mechanisms function effectively \emph{in the wild}. We define this setting through two constraints: (1) the target of detection is a benchmark rather than a massive pre-training corpus, and (2) the audited models are instruction-tuned variants that have undergone multi-stage post-training.

Our main contributions are as follows. 
\textbf{First in-the-wild evaluation of benchmark auditing.} 
We provide the first systematic evaluation of three state-of-the-art detection paradigms, \textit{LLM Dataset Inference}, \textit{Post-Hoc Dataset Inference}, and \textit{CoDeC}, beyond controlled academic corpora and in realistic benchmark auditing scenarios for which we open-source our code\footnote{\url{https://anonymous.4open.science/r/reliability-gap-benchmark-auditing/}}.

Throughout our extensive evaluation, we find that: 
\begin{enumerate}
    \item \textbf{Current auditing methods are not yet reliable enough to consistently verify benchmark exposure.}
  As summarized in ~\Cref{fig:pie_chart}, correct detection outcomes are matched by many failures across our evaluations.

    \item \textbf{At limited dataset scale, only LLM DI works, but only under favorable conditions.}
    When the investigated dataset is as small as a typical benchmark, LLM DI can still detect training exposure if it has access to a genuinely unseen and approximately IID reference set. Post-Hoc DI does not have enough data to build a reliable synthetic reference set, and CoDeC assigns similar scores to closely matched train and test splits (\Cref{sec:task_1}).

    \item \textbf{None of the methods can reliably tell which split of a benchmark was used for training.}
    In split-level auditing, LLM DI can mistake distribution differences for training exposure, Post-Hoc DI remains unstable, and CoDeC does not clearly separate seen and unseen splits from the same benchmark (\Cref{sec:task_2,sec:task_diagnosis}).
    \item \textbf{For specialised post-training datasets and industry models, we find that the signals become less stable and harder to interpret.} For domain-specific post-training datasets, detection performance degrades. For industry models, the methods can detect differences at the level of model families. 
   (\Cref{sec:task_3,sec:task_4}).
\end{enumerate}

These findings suggest several practical recommendations. LLM DI is only usable when a genuinely unseen and approximately IID validation set is available. Post-Hoc DI appears less suitable for standard benchmark-sized datasets unless the quality of the synthetic held-out data can be independently assessed. CoDeC is best interpreted as a comparative contamination indicator rather than a certification tool. Our results suggest that transparent data provenance remains the most reliable basis for benchmark-integrity claims, with statistical auditing serving as complementary evidence.

\section{Related Work}

\noindent\textbf{Membership and dataset inference.}
Membership inference attacks (MIAs) study whether a particular record was part of a model's training set, typically by exploiting systematic differences in model behavior on seen versus unseen examples~\cite{shokri2017membership}.
For large models, including LLMs, single-example MIAs can be noisy~\cite{maini2024llmdatasetinferencedid,dubinski2024towards}, motivating \emph{dataset inference} methods that aggregate weak per-sample signals over a suspect set and compare them to an unseen reference set~\cite{maini2024llmdatasetinferencedid,dubinski2025cdi,kowalczuk2025privacy}.
A key practical challenge for these tests is obtaining a reference or validation set that is truly unseen and sufficiently IID with the suspect set; violations of this assumption can lead to confounded detections.

\noindent\textbf{Benchmark contamination.}
Benchmark contamination refers to the overlap between a benchmark dataset and a model's training data~\cite{brown2020languagemodelsfewshotlearners}. This overlap may occur when benchmark train or test splits are directly included in pre-training corpora, or when near-duplicate or highly overlapping variants are present in large scraped mixtures~\cite{lee2022deduplicatingtrainingdatamakes}. In such cases, benchmark performance may reflect prior exposure rather than generalization to unseen data. As benchmarks are repeatedly reused and increasingly incorporated into post-training mixtures, contamination becomes both more likely and more difficult to detect~\cite{balloccu2024leakcheatrepeatdata}.

In parallel to documenting contamination risk, recent efforts have focused on developing methods for detecting it. Retrieval-based and behavioral auditing approaches test whether benchmark items can be recovered from candidate corpora or elicited from the model itself~\cite{deng2024investigatingdatacontaminationmodern,golchin2024timetravelllmstracing}. More recently, performance-based methods have framed contamination detection as a statistical problem, asking whether unusually strong benchmark results are better explained by prior exposure than by genuine generalization~\cite{dekoninck2024constatperformancebasedcontaminationdetection}. Another recent direction is in-context contamination detection: CoDeC~\cite{zawalski2025detectingdatacontaminationllms} measures how adding examples from the same dataset changes model confidence, using these shifts to produce an interpretable dataset-level contamination score. The method requires only gray-box access to token probabilities and assumes no prior knowledge of the training corpus, making it particularly attractive for practical benchmark auditing. 

\section{Methods}
\label{sec:methods}

In our experiments, we use the following methods: Post-Hoc Dataset Inference, LLM Dataset Inference, and CoDeC, which are described in this section.

\noindent\textbf{LLM Dataset Inference.}
This method~\cite{maini2024llmdatasetinferencedid} builds on membership inference attacks, but shifts the unit of analysis from \emph{individual samples} to \emph{datasets}. For LLMs, a single-sample MIA is often too noisy to be reliably useful: many sequences are ``easy'' (low loss) even if they were never seen during training, and the membership signal for any particular example becomes faint as models scale.

Dataset inference addresses this by \emph{aggregating weak membership signals across many samples} and testing whether, in aggregate, a \emph{suspect set} looks more train-like than an \emph{unseen validation set}. Concretely, one computes a collection of MIA-derived scores (e.g., loss or perplexity-based and related features) for samples from both sets, learns a simple aggregation of these features, and then performs a statistical test, typically a one-sided $t$-test, on held-out samples to decide whether the suspect set exhibits systematically stronger membership evidence than the validation set.

The output of dataset inference is a \textbf{$p$-value}. Under the null hypothesis $H_0$ that the model was not trained on the suspect set, the observed separation between suspect and validation should be explainable by chance. A small $p$-value (e.g., $p < 0.1$ as in~\cite{maini2024llmdatasetinferencedid}) is interpreted as evidence against $H_0$, that is, as a detection of training on the suspect dataset. Importantly, this $p$-value is only meaningful under a key assumption: the validation set must be unseen during training and IID with the suspect set. If the two sets differ in distribution, for example a benchmark train split versus its test split, the test can become confounded and may yield false positives.

\noindent\textbf{Post-Hoc Dataset Inference.}
Post-hoc DI is designed to relax the practical requirement of LLM dataset inference for an unseen, IID validation set. Instead of requiring a naturally occurring validation set, post-hoc DI constructs a synthetic validation set that is intended to be distribution-matched to the suspect data.
The procedure has two stages.

\textit{(1) Held-out data generation.} Starting from a corpus of documents for a given dataset, the method segments documents into short snippets and splits them into two disjoint pools. A small generator model is trained with a causal language modeling objective on one pool. The second pool is converted into prefix--suffix pairs; the generator is then used to produce artificial suffixes conditioned on the prefixes. The resulting dataset contains real suffixes, which form the suspect set, and synthetic suffixes, which serve as the validation set. Both sets share surface-level distributional properties, but only the suspect set could contain memorized training data from the audited model.

\textit{(2) Post-hoc calibration.} The method trains two classifiers to distinguish suspect versus synthetic validation examples: (i) a text-only classifier $c_{\text{text}}$, and (ii) a combined classifier $c_{\text{comb}}$ that augments text with MIA-derived features such as loss statistics. The central statistical question is whether the MIA features provide additional separability beyond what is already achievable from text alone. Concretely, one computes per-example classification scores and performs a one-sided $t$-test to determine whether the combined classifier improves separation relative to the text-only baseline. The output is a \textbf{$p$-value} under the null hypothesis $H_0$ that MIA features do not add meaningful signal once text is accounted for. A small $p$-value, for example $p < 0.1$, is interpreted as evidence that the audited model was trained on the suspect distribution.

\noindent\textbf{Context-based contamination signals.}
Complementary to likelihood- and classifier-based auditing, CoDeC (Contamination Detection via Context) detects dataset-level contamination by measuring how same-dataset in-context examples affect next-token likelihood~\cite{zawalski2025detectingdatacontaminationllms}. In-context learning often improves predictions on unseen data, but can become unhelpful or harmful when memorization is triggered. This yields an interpretable contamination score defined as the fraction of samples for which in-context examples reduce model confidence.

The key observation is that in-context learning is typically helpful when the model has not internalized the target distribution. Adding a few examples provides dataset-specific cues such as format or vocabulary and improves next-token likelihood. In contrast, if the model has already been trained on the dataset or a close variant, in-context examples provide little new information.

Operationally, for each sample $x$ from a dataset $\mathcal{D}$, CoDeC compares the model's average log-likelihood on $x$ in two settings: (i) a baseline setting where the model predicts $x$ directly, and (ii) an in-context setting where $x$ is preceded by a small number of other samples from $\mathcal{D}$. The per-sample score is the log-likelihood difference $\Delta(x)$, defined as in-context minus baseline. The dataset \textbf{contamination score} is the fraction of samples for which $\Delta(x) < 0$. Higher values indicate that in-context learning is frequently unhelpful or harmful, which is consistent with stronger contamination.

\section{Experimental setup}

In this section, we describe the experimental setup used to evaluate contamination-detection methods across multiple model families, datasets, and auditing scenarios. We create a publicly available \href{https://anonymous.4open.science/r/reliability-gap-benchmark-auditing/README.md}{codebase} which adapts each method to shared interface for benchmarking and detail the method configurations, audited models and benchmarks, and the notation used to interpret detection outcomes in the experiments that follow.

\subsection{Detection Methods Configuration.}
    \noindent\textbf{LLM DI:} This method mirrors the MIA metrics, statistical test, outlier handling, and normalization strategy of Post-Hoc DI. Datasets are evenly split: set~A computes metrics and trains a linear classifier using 250 samples per side, while set~B evaluates the final $t$-test using at least 1\,000 texts, the maximum number of texts per split evaluated by authors of methods~\cite{maini2024llmdatasetinferencedid}.

    \noindent\textbf{Post-Hoc DI:} We follow the original method as closely as possible in our benchmark-auditing setting. We subsample 2\,000 texts per set and clip outliers symmetrically at 2.5\% per tail. A \texttt{Meta-Llama-3.1-8B} model fine-tuned with LoRA generates 128-token suffixes for paraphrasing. Calibration relies on a GPT-2 baseline text classifier, while combined features are evaluated using a linear classifier trained for 1\,000 epochs with Adam. MIA metrics are $z$-score normalized prior to training. We utilize perplexity, $k$-min-probs ($k{=}0.05$), reversed $k$-max-probs ($k{=}0.05$), and zlib ratio as our MIA metrics. We apply a one-sided independent samples $t$-test with a significance threshold of $p < 0.1$,\footnote{increased from $p < 0.05$ due to  high p-values for members (where an ideal method should have very low p-values for members) and consistency with LLM DI} taking the mean over 3 independent runs.
    
    \noindent\textbf{CoDeC:} The contamination score is calculated as a binary per-sample classification, assigning a value of 1 if confidence drops when context is added. To avoid prompt artifacts, log-probabilities are averaged from the 10th token onward, we strictly utilize 1 randomly sampled context example --- following the official description of CoDeC pipeline~\cite{zawalski2025detectingdatacontaminationllms}.

\subsection{Audited Models and Datasets.}

    \noindent\textbf{Open Model Suites:} Fully open model suites enable rigorous contamination studies by allowing inspection of training data to support ground-truth membership claims. We evaluate the \textbf{Pythia suite}, known for its tight coupling with \textbf{The Pile} reference corpus. We also audit \textbf{OLMo 2}, which details its broad knowledge pre-training~\cite{olmo20252olmo2furious}, rigorous post-training (incorporating datasets like \textit{UltraFeedback}~\cite{cui2023ultrafeedback}, \textit{tulu3 sft math}~\cite{Lambert2024TLU3P}, \textit{Competition Math}~\cite{hendrycks2021math}, and \textit{GSM8K}~\cite{cobbe2021gsm8k}), and extensive evaluation phases.
    
    \noindent\textbf{Evaluation Benchmarks:} For the open-model experiments, we evaluate benchmark exposure on OLMo~2 using high-impact evaluation sets frequently audited for leakage as their train splits are widely available and exert strong optimization pressure. Representative benchmarks include \textbf{GSM8K}~\cite{cobbe2021gsm8k} (grade-school math word problems), \textbf{MATH} and \textbf{Competition Math}~\cite{hendrycks2021math} (multi-step competition mathematics), \textbf{DROP}~\cite{dua2019drop} (discrete reasoning over paragraphs), and \textbf{MMLU}~\cite{hendrycks2021mmlu} (multi-domain knowledge evaluation).
    
    \noindent\textbf{PLLuM:} A family of Polish language LLMs~\cite{kocoń2025pllumfamilypolishlarge}, which undergoes supervised fine-tuning using automatic and manual datasets, followed by preference-based alignment. We audit the Llama-PLLuM-8B, PLLuM-12B, and non-commercial PLLuM-12B-nc variants.
    
    \noindent\textbf{Medical LLMs:} We evaluate domain-specialised models fine-tuned on public medical QA benchmarks. Tested benchmarks include MedQA-USMLE~\cite{jin2020disease}, MedMCQA~\cite{pal2022medmcqalargescalemultisubject}, PubMedQA~\cite{jin-etal-2019-pubmedqa}, and MedExpQA~\cite{alonso2024medexpqamultilingualbenchmarkinglarge}. Audited models include variants of MedGemma~\cite{sellergren2025medgemmatechnicalreport} (4B, 27B), Meditron3~\cite{sallinen2025llamameditron} (8B, 9B), Meerkat~\cite{kim2024smalllanguagemodelslearn} (7B, 8B), and Neeto-1.0-8B~\cite{Neeto-1.0-8b}.

\subsection{Notation for Detection Outcomes.}
Throughout the tables in this section, we report detection outcomes using colored symbols to indicate correctness at a glance:
\begin{itemize}[label=\textbullet]
    \item \gplus{} (green~\textbf{+}): \emph{true positive}; the method correctly identifies that the model was trained on the given data.
    \item \gminus{} (green~\textbf{--}): \emph{true negative}; the method correctly determines that the model was not trained on the given data.
    \item \rplus{} (red~\textbf{+}): \emph{false positive}; the method incorrectly indicates that the model was trained on the data when it was not.
    \item \rminus{} (red~\textbf{--}): \emph{false negative}; the method fails to detect that the model was in fact trained on the data.
\end{itemize}

The detection decision (\textbf{+} vs.\ \textbf{--}) is derived differently for each method. 
For \emph{LLM Dataset Inference}, a detection (\textbf{+}) is declared when the one-sided $t$-test $p$-value falls below~$0.1$, following the threshold used in~\cite{maini2024llmdatasetinferencedid}. 
For \emph{CoDeC}, a detection (\textbf{+}) is declared when the contamination score is higher than $0.8$ indicating strong contamination, no detection (\textbf{--}) when contamination score is below $0.6$, and inconclusive (\textbf{?}) when the score is between $0.6$ and $0.8$, as suggested in the original work~\cite{zawalski2025detectingdatacontaminationllms}. Tables with numerical values instead of symbols can be found in the supplementary material.

\section{Experiments}
    
We first test whether current contamination detection methods remain effective when only benchmark-scale reference sets are available for inspection, rather than full pre-training corpora (\Cref{sec:task_1}). We then examine whether these methods can distinguish between seen and unseen splits within the same benchmark (\Cref{sec:task_2}). Next, we evaluate the most effective approaches on specialised post-training datasets (\Cref{sec:task_3}). Finally, we apply  a comparative analysis on industry models (\Cref{sec:task_4}) and analyze the failure modes underlying these results (\Cref{sec:task_diagnosis}). 

\subsection{Task 1: Auditing with Limited Reference Data}
\label{sec:task_1}

A fundamental challenge in applying current detection methods ``in the wild'' is the vast disparity in data scale: while pre-training corpora typically encompass gigabytes of text, specific evaluation benchmarks often consist of only a few megabytes. To rigorously evaluate detection performance under these constrained conditions, we simulate the data scarcity characteristic of benchmark auditing. We find that Post-Hoc DI becomes ineffective when trained on only a few thousand examples, while CoDeC cannot distinguish split-level membership when train and test distributions are too similar. Only LLM DI, which requires access to an IID validation set, remains reliable at benchmark scale.

\textbf{Task Setup.}
Instead of utilizing the full corpus, we restrict the reference data to $n=2,000$ documents per dataset subsampled from various subsets of The Pile. We then evaluate the detection methods across multiple Pythia model sizes, effectively mimicking a scenario where the auditor has access to only a fraction of the data distribution, comparable to the size of a standard evaluation benchmark, rather than the massive datasets used in prior academic validations.

\begin{table}[h!]
\caption{\textbf{Detection of training on train and test splits of the Pile (Limited Data).} We reorganize the results into a grid to highlight consistency across diverse subsets. Symbols: \gplus\ = correctly detected as trained on, \gminus\ = correctly detected as \emph{not} trained on, \rplus\ = false positive, \rminus\ = false negative, \oquest\ = uncertain signal.}
\label{tab:task1_results}
\centering
\resizebox{\textwidth}{!}{%
\begin{tabular}{lccccc}
\toprule
& \multicolumn{2}{c}{Pile CC} & \multicolumn{2}{c}{Pile Europarl} & Corr./Tot. \\
\cmidrule(lr){2-3} \cmidrule(lr){4-5}
& Train & Test & Train & Test & \\
\textit{Pythia sizes} 
& \multicolumn{1}{c}{0.4B/1.4B/2.8B/6.9B} 
& \multicolumn{1}{c}{0.4B/1.4B/2.8B/6.9B} 
& \multicolumn{1}{c}{0.4B/1.4B/2.8B/6.9B} 
& \multicolumn{1}{c}{0.4B/1.4B/2.8B/6.9B} 
& \\
\midrule
LLM DI 
& \rminus / \rminus / \rminus / \rminus 
& \gminus / \gminus / \gminus / \gminus 
& \gplus / \gplus / \gplus / \gplus 
& \gminus / \gminus / \gminus / \gminus 
& 12/16 \\
Post-Hoc DI 
& \rminus / \rminus / \rminus / \rminus 
& \gminus / \gminus / \gminus / \gminus 
& \rminus / \rminus / \rminus / \rminus 
& \gminus / \gminus / \gminus / \gminus 
& 8/16 \\
CoDeC 
& \gplus / \gplus / \gplus / \gplus 
& \rplus / \rplus / \rplus / \rplus 
& \gplus / \gplus / \gplus / \gplus 
& \rplus / \rplus / \rplus / \rplus 
& 8/16 \\
\midrule
\midrule
& \multicolumn{2}{c}{Pile Hacker News} & \multicolumn{2}{c}{Pile Stack Exchange} & Corr./Tot. \\
\cmidrule(lr){2-3} \cmidrule(lr){4-5}
& Train & Test & Train & Test & \\
\textit{Pythia sizes} 
& \multicolumn{1}{c}{0.4B/1.4B/2.8B/6.9B} 
& \multicolumn{1}{c}{0.4B/1.4B/2.8B/6.9B} 
& \multicolumn{1}{c}{0.4B/1.4B/2.8B/6.9B} 
& \multicolumn{1}{c}{0.4B/1.4B/2.8B/6.9B} 
& \\
\midrule
LLM DI
& \gplus / \gplus / \gplus / \gplus
& \gminus / \gminus / \gminus / \gminus
& \gplus / \gplus / \gplus / \gplus
& \gminus / \gminus / \gminus / \gminus
& 16/16 \\
Post-Hoc DI 
& \rminus / \rminus / \rminus / \rminus 
& \rplus / \rplus / \rplus / \rplus 
& \rminus / \rminus / \rminus / \rminus 
& \gminus / \gminus / \gminus / \gminus 
& 4/16 \\
CoDeC 
& \gplus / \gplus / \gplus / \gplus 
& \rplus / \rplus / \rplus / \rplus 
& \gplus / \gplus / \gplus / \gplus 
& \rplus / \rplus / \rplus / \rplus 
& 8/16 \\
\bottomrule
\end{tabular}%
}
\end{table}

\textbf{Findings.} As shown in \Cref{tab:task1_results}, LLM Dataset Inference correctly distinguishes train from test splits across all model sizes, confirming that our implementation reproduces prior academic findings under controlled conditions. 
Post-Hoc Dataset Inference, however, does not reject the null hypothesis of non-membership in most cases. At benchmark scale, the generator trained on only 2{,}000 samples per subset is too weak to produce realistic synthetic validation data, leading to uninformative calibration statistics.
This interpretation is supported by an analysis of a full-scale held-out generation setting at~\Cref{sec:task_diagnosis}.
CoDeC assigns high contamination scores to both splits when distributions are similar. As a result, high scores may reflect exposure to either the train split alone or both splits, indicating sensitivity to distributional overlap rather than precise membership.

\subsection{Task 2: Detecting Split-Level Benchmark Exposure}
\label{sec:task_2}

In this task, we move from controlled Pythia/Pile simulations to realistic benchmark auditing on instruction-tuned OLMo~2, and assess if current methods can distinguish between seen and unseen splits within the same benchmark. We find that none of the methods can robustly distinguish seen from unseen splits once the setting shifts from corpus-level datasets to split-level benchmark exposure.

\textbf{Task Setup.}
We utilize OLMo~2, whose post-training mixture explicitly includes the training split of GSM8K, while their test splits are reserved for evaluation. As controls for entirely unseen benchmark data, we include MMLU and DROP: although OLMo~2 is evaluated on these benchmarks, it was not trained on their training splits. Our objective is to verify if detection mechanisms can correctly flag the train splits of GSM8K and Competition Math as contaminated, while refraining from flagging their test splits or the entirely held-out benchmarks (MMLU and DROP).

\begin{table}[H]
\caption{\textbf{Detection of training on train and test splits for OLMo~2.} Symbols: \gplus\ = correctly detected as trained on, \gminus\ = correctly detected as \emph{not} trained on, \rplus\ = falsely detected as trained on (false positive), \rminus\ = missed detection (false negative), \oquest\ = uncertain signal. Test set of GSM8K was not large enough to run Post-Hoc DI.}
\label{tab:task2_results}
\centering
\resizebox{1\textwidth}{!}{%
\begin{tabular}{lccccccc}
\toprule
& \multicolumn{2}{c}{GSM8K} & \multicolumn{2}{c}{MMLU} & \multicolumn{2}{c}{DROP} & Corr./Tot. \\
\cmidrule(lr){2-3} \cmidrule(lr){4-5} \cmidrule(lr){6-7}
& Train & Test & Train & Test & Train & Test & \\
\textit{OLMo 2 version} & \multicolumn{2}{c}{1B / 7B / 13B} & \multicolumn{2}{c}{1B / 7B / 13B} & \multicolumn{2}{c}{1B / 7B / 13B} & \\
\midrule
LLM DI 
& \gplus / \gplus / \gplus 
& \gminus / \gminus / \gminus 
& \rplus / \rplus / \rplus
& \gminus / \gminus / \gminus
& \rplus / \rplus / \rplus
& \gminus / \gminus / \gminus
& 12/18 \\
Post-Hoc DI
& \rminus / \rminus / \gplus
& n/a / n/a / n/a
& \gminus / \gminus / \gminus
& \gminus / \gminus / \gminus
& \gminus / \gminus / \gminus
& \gminus / \gminus / \gminus
& 13/18 \\
CoDeC 
& \gplus / \rminus / \rminus 
& \rplus / \rplus / \oquest 
& \gminus / \oquest / \gminus
& \gminus / \gminus / \gminus
& \rplus / \gminus / \gminus
& \rplus / \oquest / \gminus
& 9/18 \\
\bottomrule
\end{tabular}%
}
\end{table}

\begin{figure}[!h]
    \centering
    \includegraphics[width=1\linewidth]{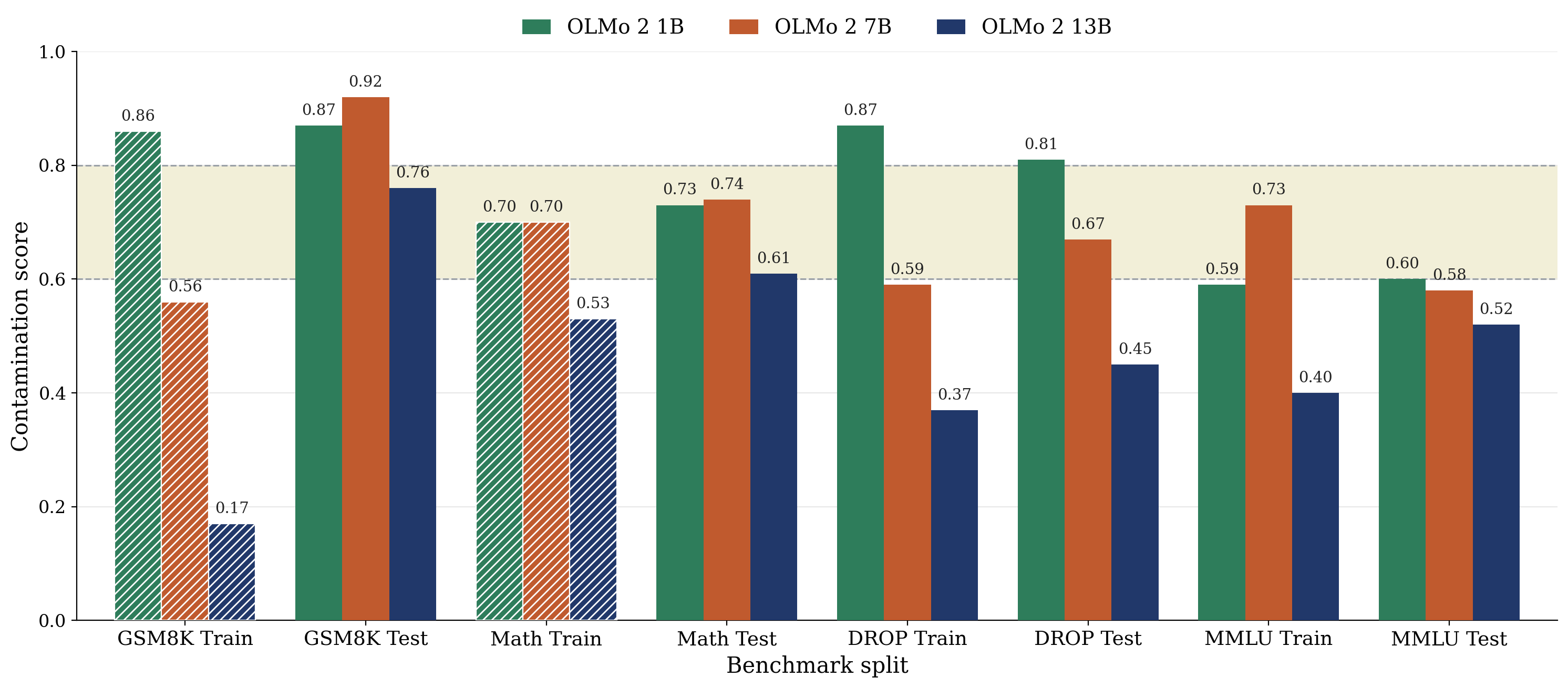}
    \caption{\textbf{CoDeC scores for Task 2}. CoDeC contamination scores for benchmarks. Hatched bars indicate that this split was used as training split. We can observe trend between model sizes, however training splits do not have higher scores.}
    \label{fig:task2_bar_codec}
\end{figure}

\textbf{Findings.} The results in \Cref{tab:task2_results} reveal clear failure modes for each method. LLM Dataset Inference produces false positives on DROP and MMLU even though OLMo~2 was not trained on their train splits. One possible explanation is that, although train and test splits of the same benchmark appear to be natural IID proxies, they can still differ in difficulty, style, or construction. In such cases, LLM DI can be driven by split-level distribution differences rather than by training exposure.
Post-Hoc Dataset Inference again fails to cleanly separate train and test splits. As in Task 1, synthetic validation data introduces artifacts that dominate the statistical test. 
CoDeC assigns elevated contamination scores to GSM8K, consistent with its known post-training exposure; however, DROP and MMLU receive only slightly lower scores, so the signal does not provide clear evidence of split-level membership. Instead, the scores decrease with model size, a trend that \Cref{fig:task2_bar_codec} reinforces by extending the same pattern to a further member benchmark, Competition Math.

\subsection{Task 3: Auditing Specialised Post-Training Datasets}
\label{sec:task_3}

In this task, we examine whether contamination detection transfers from controlled benchmark settings to specialised post-training domains, specifically medical LLMs and mid-resource language models. We find that detection performance is highly inconsistent across these settings and deteriorates substantially when the available benchmark splits are small.

\textbf{Task Setup.}
We apply LLM DI and CoDeC, the two methods that remained most effective in earlier experiments and were also the most practical to apply, to two downstream settings. For medical question answering, we evaluate MedGemma~\cite{sellergren2025medgemmatechnicalreport}, Meditron3~\cite{sallinen2025llamameditron}, Meerkat~\cite{kim2024smalllanguagemodelslearn}, and Neeto-1.0~\cite{Neeto-1.0-8b} on the train and test splits of MedQA-USMLE~\cite{jin2020disease} and MedMCQA~\cite{pal2022medmcqalargescalemultisubject}. For Polish culture-aware language models, we evaluate the PLLuM-12B and Llama-PLLuM-8B families on train/test (or train/validation) splits from Automatic SFT, Manual SFT, and Alignment datasets. Because some held-out splits are very small, we subsample the larger split to match the smaller one, yielding evaluation sets of approximately 1{,}600 examples for Automatic SFT and 1{,}150 for Manual SFT. This setting tests whether contamination signals remain informative on small, domain-specific post-training datasets.

\begin{table}[h]
\centering
\caption{\textbf{Detection of training on train and test splits for medical LLMs.} Symbols: \gplus\ = correctly detected as trained on, \gminus\ = correctly detected as \emph{not} trained on, \rplus\ = falsely detected as trained on (false positive), \rminus\ = missed detection (false negative), \oquest\ = uncertain signal.}
\label{tab:medical-contamination}
\scriptsize
\begin{tabular}
{@{\extracolsep{\fill}}ll|cc|cc|c}
\toprule
& & \multicolumn{2}{c|}{\textbf{MedQA-USMLE}} & \multicolumn{2}{c|}{\textbf{MedMCQA}} & \multirow{2}{*}{\textbf{Corr./Tot.}} \\
\cmidrule(lr){3-4} \cmidrule(lr){5-6}
& & Train & Test & Train & Test & \\
\midrule
\multirow{2}{*}{\makecell{medgemma\\4B / 27B}}
& LLM DI & \rminus / \rminus & \gminus / \gminus & \rminus / \gplus & \gminus / \gminus & 5/8 \\
& CoDeC  & \rminus / \rminus & \gminus / \gminus & \rminus / \rminus & \gminus / \gminus & 4/8 \\
\midrule
\multirow{2}{*}{\makecell{Meditron3\\8B / 9B}} 
& LLM DI & \rminus / \rminus & \gminus / \gminus & \gplus / \gplus & \gminus / \gminus & 6/8 \\
& CoDeC  & \oquest / \rminus & \oquest / \gminus & \rminus / \rminus & \gminus / \gminus & 3/8 \\
\midrule
\multirow{2}{*}{\makecell{meerkat\\7B / 8B}}
& LLM DI & \rminus / \rminus & \gminus / \gminus & \gplus / \rminus & \gminus / \gminus & 5/8 \\
& CoDeC  & \oquest / \oquest & \oquest / \oquest & \rminus / \rminus & \gminus / \gminus & 2/8 \\
\midrule
\multirow{2}{*}{\makecell{Neeto-1.0\\8B}} 
& LLM DI & \rminus & \gminus & \rminus & \gminus & 2/4 \\
& CoDeC  & \rminus & \gminus & \rminus & \gminus & 2/4 \\
\bottomrule
\end{tabular}
\end{table}

\begin{table}[h]
\centering
\caption{\textbf{Detection of training on SFT and alignment data for PLLuM models.} Symbols: \gplus\ = correctly detected as trained on, \gminus\ = correctly detected as \emph{not} trained on, \rplus\ = falsely detected as trained on (false positive), \rminus\ = missed detection (false negative), \oquest\ = uncertain signal. base* = updated base checkpoint (250801).}
\label{tab:pllum-sft-contamination}
\footnotesize
\resizebox{\linewidth}{!}{%
\begin{tabular}{ll|cc|cc|cc|c}
\toprule
& & \multicolumn{2}{c|}{\textbf{Automatic SFT}} & \multicolumn{2}{c|}{\textbf{Manual SFT}} & \multicolumn{2}{c|}{\textbf{Alignment}} & \textbf{Corr./Tot.} \\
\cmidrule(lr){3-4} \cmidrule(lr){5-6} \cmidrule(lr){7-8}
& & Train & Val & Train & Val & Train & Test & \\
\midrule
\multirow{3}{*}{\makecell{PLLuM-12B\\base}} & \textit{base / base* / nc-base} &  &  &  &  &  &  &  \\
 & LLM DI & \gminus / \gminus / \gminus & \gminus / \gminus / \gminus & \gminus / \gminus / \gminus & \gminus / \gminus / \gminus & \rplus / \rplus / \rplus & \gminus / \gminus / \gminus & 15/18 \\
 & CoDeC & \rplus / \gminus / \gminus & \rplus / \gminus / \gminus & \oquest / \gminus / \gminus & \oquest / \gminus / \gminus & \gminus / \gminus / \gminus & \gminus / \gminus / \gminus & 14/18 \\
\midrule
\multirow{3}{*}{\makecell{PLLuM-12B\\SFT}} & \textit{chat / inst. / nc-chat / nc-inst.} &  &  &  &  &  &  &  \\
 & LLM DI & \gplus / \rminus / \rminus / \rminus & \gminus / \gminus / \gminus / \gminus & \rminus / \rminus / \rminus / \rminus & \gminus / \gminus / \gminus / \gminus & \gplus / \rplus / \gplus / \rplus & \gminus / \gminus / \gminus / \gminus & 15/24 \\
 & CoDeC & \gplus / \gplus / \gplus / \oquest & \rplus / \rplus / \oquest / \gminus & \oquest / \oquest / \oquest / \rminus & \oquest / \oquest / \rplus / \oquest & \oquest / \gminus / \gplus / \gminus & \oquest / \gminus / \rplus / \gminus & 9/24 \\
\midrule
\multirow{3}{*}{\makecell{Llama-PLLuM\\8B base}} & \textit{base / base*} &  &  &  &  &  &  &  \\
 & LLM DI & \gminus / \gminus & \gminus / \gminus & \gminus / \gminus & \gminus / \gminus & \rplus / \rplus & \gminus / \gminus & 10/12 \\
 & CoDeC & \rplus / \gminus & \rplus / \gminus & \oquest / \oquest & \rplus / \gminus & \gminus / \gminus & \gminus / \gminus & 7/12 \\
\midrule
\multirow{3}{*}{\makecell{Llama-PLLuM\\8B SFT}} & \textit{chat / inst.} &  &  &  &  &  &  &  \\
 & LLM DI & \rminus / \rminus & \gminus / \gminus & \rminus / \rminus & \gminus / \gminus & \gplus / \rplus & \gminus / \gminus & 7/12 \\
 & CoDeC & \gplus / \gplus & \rplus / \rplus & \gplus / \oquest & \rplus / \rplus & \gplus / \gminus & \rplus / \oquest & 5/12 \\
\bottomrule
\end{tabular}%
}
\end{table}

\textbf{Findings.}
The results in \Cref{tab:medical-contamination,tab:pllum-sft-contamination} show that detection performance degrades substantially in downstream settings. For medical QA, LLM DI detects training on the MedMCQA train split for most models, though it misses it for medgemma-4B, meerkat-8B, and Neeto-1.0, and it consistently fails on MedQA-USMLE, missing the train split for every model. Held-out test splits are uniformly identified as non-members. CoDeC is less stable: it correctly identifies some train splits, but its behavior varies across models and does not consistently separate seen from unseen data.

For PLLuM, both methods are further limited by the small size of the evaluation splits. LLM DI often misses training on Automatic and Manual SFT, yielding repeated false negatives on train splits, while also producing false positives on Alignment. CoDeC shows partial success for the PLLuM-12B family, but performs substantially worse for Llama-PLLuM-8B and again tends to assign elevated scores to both train and held-out splits from the same dataset.

\subsection{Task 4: Comparative Auditing of Industry Models}
\label{sec:task_4}

\begin{figure}[h!]
    \centering
    \includegraphics[width=0.95\linewidth]{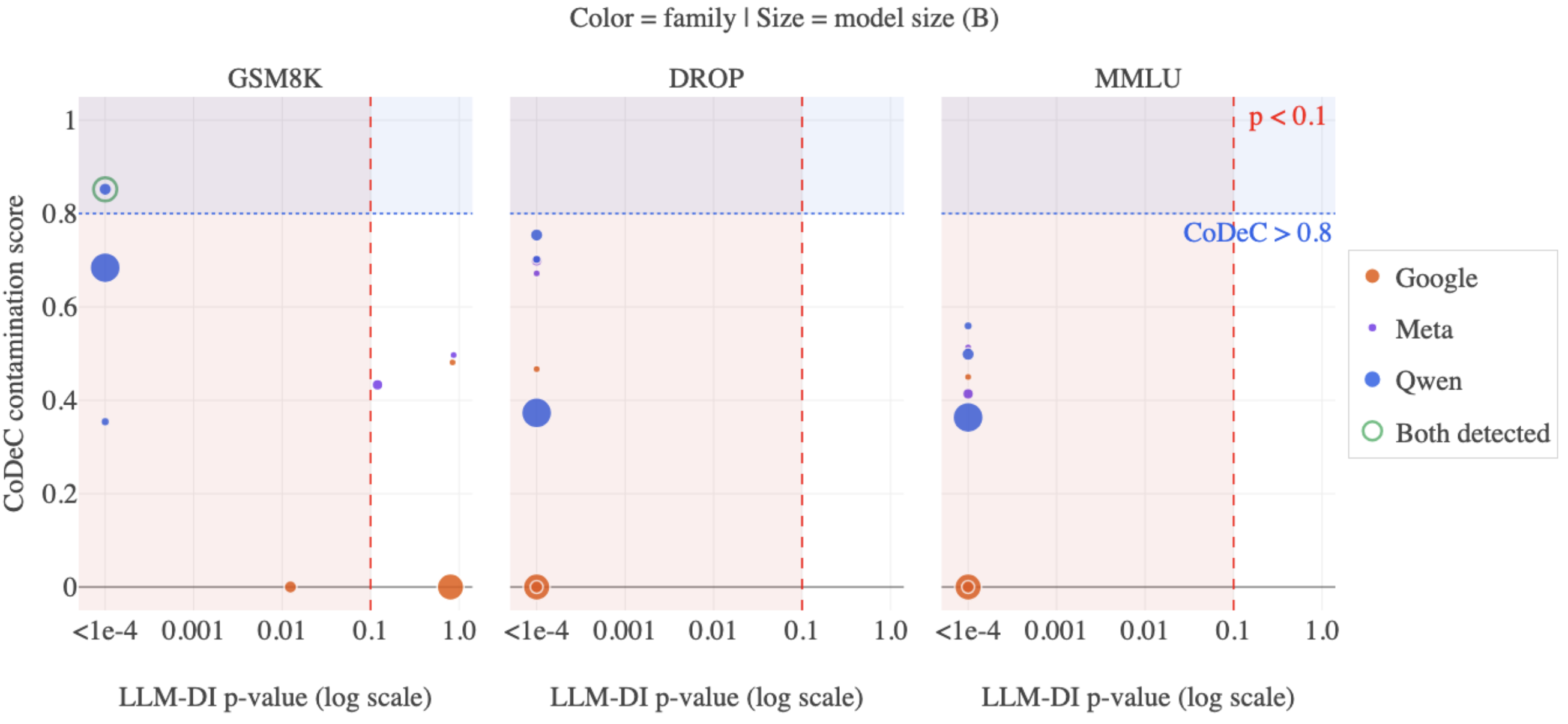}
    \caption{\textbf{Application of CoDeC and LLM Dataset Inference to industry models.} We try to detect training on train splits of frontier benchmarks.}
    \label{fig:industry}
\end{figure}

In this task, we extend the analysis to closed or partially disclosed industry models. We ask whether LLM DI and CoDeC provide consistent signals of prior exposure to benchmark data for frontier evaluations. We find that the signals are suggestive rather than definitive: most model families are flagged on at least some benchmarks, while Google's Gemma-3 family stands out as the only series not flagged by LLM DI on GSM8K and having consistently lower CoDeC scores.

\textbf{Task Setup.}
We evaluate LLM Dataset Inference and CoDeC on several industry model families --- Google: Gemma 3 (1B, 4B, 12B); Alibaba: Qwen 3 (1.7B, 4B, 14B); Meta: Llama 3.2 (1B, 3B) --- using public benchmarks: GSM8K, MMLU, and DROP. Because training data for these models is not fully disclosed, ground-truth membership is unavailable. Our goal is therefore comparative: to assess whether contamination signals differ systematically across model families.

\textbf{Findings.}
As shown in \Cref{fig:industry}, most model families receive elevated CoDeC scores across benchmarks, and LLM DI frequently flags benchmark training. Gemma-3 is the main exception: it consistently shows lower CoDeC scores, and on GSM8K it is the only family not flagged by LLM DI on the train split. Based on this evidence, we infer that the Gemma-3 series was not likely trained on the GSM8K training set.

\subsection{Diagnosis of Failure}
\label{sec:task_diagnosis}

The methods evaluated in previous tasks do not yield consistent outcomes.
As shown in ~\Cref{fig:pie_chart}, the number of successful detections is comparable to the number of failures; overall, just over 60\% of the outcomes are correct. To understand these results, we examine the assumptions and failure modes of each method.

\paragraph{LLM Dataset Inference.} LLM DI critically depends on the availability of a validation set that is both unseen and truly IID with the suspect dataset. This explains the contrast between Tasks~1 and later Tasks: LLM DI works in the controlled setting of Task~1. However, benchmark splits used in the later Tasks may break the IID assumption. These splits may differ in difficulty or construction, leading to the test capturing such differences rather than training exposure.

\paragraph{Post-Hoc Dataset Inference.} \label{posthoc-diagnosis}
On benchmark-sized subsets, Post-Hoc DI is dominated by the mismatch between real and synthetic held-out text rather than by memorization signal. As shown in \Cref{tab:posthoc-aucs-subsample}, the text-only classifier is already highly discriminative, indicating strong separability between real and synthetic suffixes even without MIA features.
In a well-calibrated Post-Hoc DI setup, MIA features should provide an additional discriminative signal beyond the text-only baseline. Here, however, adding MIA features yields only marginal improvements. For example, Common Crawl (410M) increases from $0.880$ to $0.899$, and Europarl (Train) slightly decreases from $0.895$ to $0.890$. This shows that the classification task is driven primarily by real-versus-synthetic distribution shift, not by training membership signal.

% Table 5
\begin{table}[b]
\centering
\caption{\textbf{Calibration-stage classifier AUCs, benchmark-scale subsampling.} \textit{Base} uses text only; \textit{Comb} adds MIA features. Uniformly high Base AUCs ($>0.78$) mean the real-vs-synthetic distribution shift alone is discriminative and MIA features barely help---the regime in which Post-Hoc DI fails. Cf.\ full-scale \Cref{tab:posthoc-aucs-small-pile}.}
\label{tab:posthoc-aucs-subsample}
% \resizebox{1\linewidth}{!}{%
\scriptsize
\begin{tabular}{l|cc|cc|cc|cc}
\toprule
& \multicolumn{2}{c|}{\textbf{Pythia 410M}} & \multicolumn{2}{c|}{\textbf{Pythia 1.4B}} & \multicolumn{2}{c|}{\textbf{Pythia 2.8B}} & \multicolumn{2}{c}{\textbf{Pythia 6.9B}} \\
\cmidrule(lr){2-3} \cmidrule(lr){4-5} \cmidrule(lr){6-7} \cmidrule(lr){8-9}
Dataset (Split) & \textit{Base} & \textit{Comb} & \textit{Base} & \textit{Comb} & \textit{Base} & \textit{Comb} & \textit{Base} & \textit{Comb} \\
\midrule
CC (Train) & 0.880 & 0.899 & 0.880 & 0.901 & 0.880 & 0.902 & 0.880 & 0.901 \\
CC (Val) & 0.882 & 0.899 & 0.882 & 0.900 & 0.882 & 0.902 & 0.882 & 0.900 \\
\midrule
Europarl (Train) & 0.895 & 0.890 & 0.895 & 0.893 & 0.895 & 0.894 & 0.895 & 0.895 \\
Europarl (Val) & 0.887 & 0.890 & 0.887 & 0.891 & 0.887 & 0.893 & 0.887 & 0.893 \\
\midrule
Hacker News (Train) & 0.899 & 0.911 & 0.899 & 0.912 & 0.899 & 0.915 & 0.899 & 0.914 \\
Hacker News (Val) & 0.899 & 0.928 & 0.899 & 0.929 & 0.899 & 0.929 & 0.899 & 0.930 \\
\midrule
Stack Exchange (Train) & 0.788 & 0.804 & 0.788 & 0.803 & 0.788 & 0.805 & 0.788 & 0.803 \\
Stack Exchange (Val)   & 0.832 & 0.844 & 0.832 & 0.844 & 0.832 & 0.845 & 0.832 & 0.844 \\
\bottomrule
\end{tabular}
\end{table}

% Table 6
\begin{table}[h]
\centering
\caption{\textbf{Calibration-test $p$-values, benchmark-scale subsampling ($n=2{,}000$).} Symbols: \gplus\ = correctly trained on ($p < 0.1$), \gminus\ = correctly \emph{not} trained on, \rplus\ = false positive, \rminus\ = false negative. The test fails to reject the null for nearly every train split (12/32 correct), the failure mode caused by the distribution shift in \Cref{tab:posthoc-aucs-subsample}. Cf.\ full-scale \Cref{tab:posthoc-p-values-new-mias}.}
\label{tab:posthoc-pvalues-subsample}
\scriptsize
\resizebox{\linewidth}{!}{%
\begin{tabular}{@{\extracolsep{\fill}}l|cc|cc|cc|cc|c}
\toprule
& \multicolumn{2}{c|}{\textbf{Pythia 410M}} & \multicolumn{2}{c|}{\textbf{Pythia 1.4B}} & \multicolumn{2}{c|}{\textbf{Pythia 2.8B}} & \multicolumn{2}{c|}{\textbf{Pythia 6.9B}} & \textbf{Corr./Tot.} \\
\cmidrule(lr){2-3} \cmidrule(lr){4-5} \cmidrule(lr){6-7} \cmidrule(lr){8-9}
Dataset & Train & Val & Train & Val & Train & Val & Train & Val & \\
\midrule
Pile CC 
& $0.583$ \rminus & $0.396$ \gminus
& $0.444$ \rminus & $0.281$ \gminus
& $0.402$ \rminus & $0.258$ \gminus
& $0.385$ \rminus & $0.236$ \gminus
& 4/8 \\
Europarl 
& $1.000$ \rminus & $1.000$ \gminus
& $1.000$ \rminus & $0.999$ \gminus
& $1.000$ \rminus & $0.999$ \gminus
& $1.000$ \rminus & $0.999$ \gminus
& 4/8 \\
Hacker News 
& $0.450$ \rminus & $0.049$ \rplus
& $0.441$ \rminus & $0.007$ \rplus
& $0.315$ \rminus & $0.006$ \rplus
& $0.388$ \rminus & $0.009$ \rplus
& 0/8 \\
Stack Exch. 
& $0.709$ \rminus & $0.821$ \gminus
& $0.707$ \rminus & $0.811$ \gminus
& $0.700$ \rminus & $0.755$ \gminus
& $0.747$ \rminus & $0.788$ \gminus
& 4/8 \\
\bottomrule
\end{tabular}%
}
\end{table}

% Table 7
\begin{table}[h]
\centering
\caption{\textbf{Calibration-stage classifier AUCs, full-scale held-out generation} (authors' released sets, generator trained on nearly the full subset). \textit{Base} uses text only; \textit{Comb} adds MIA features. Base AUCs fall to around $55\%$ (e.g.\ Common Crawl Base: $0.548$): real and synthetic suffixes become nearly indistinguishable and the distribution-shift artifact of \Cref{tab:posthoc-aucs-subsample} largely vanishes.}
\label{tab:posthoc-aucs-small-pile}
\scriptsize
\setlength{\tabcolsep}{3pt}
\begin{tabular}{l|cc|cc|cc|cc}
\toprule
& \multicolumn{2}{c|}{\textbf{Pythia 410M}} & \multicolumn{2}{c|}{\textbf{Pythia 1.4B}} & \multicolumn{2}{c|}{\textbf{Pythia 2.8B}} & \multicolumn{2}{c}{\textbf{Pythia 6.9B}} \\
\cmidrule(lr){2-3} \cmidrule(lr){4-5} \cmidrule(lr){6-7} \cmidrule(lr){8-9}
Dataset (Split) & \textit{Base} & \textit{Comb} & \textit{Base} & \textit{Comb} & \textit{Base} & \textit{Comb} & \textit{Base} & \textit{Comb} \\
\midrule
CC (Train) & 0.548 & 0.557 & 0.548 & 0.556 & 0.548 & 0.564 & 0.548 & 0.567 \\
CC (Val) & 0.506 & 0.513 & 0.506 & 0.515 & 0.506 & 0.517 & 0.506 & 0.523 \\
\midrule
Europarl (Train) & 0.509 & 0.532 & 0.509 & 0.555 & 0.509 & 0.567 & 0.509 & 0.575 \\
Europarl (Val) & 0.514 & 0.516 & 0.514 & 0.523 & 0.514 & 0.526 & 0.514 & 0.533 \\
\midrule
Hacker News (Train) & 0.525 & 0.567 & 0.525 & 0.578 & 0.525 & 0.585 & 0.525 & 0.591 \\
Hacker News (Val) & 0.570 & 0.569 & 0.570 & 0.571 & 0.570 & 0.578 & 0.570 & 0.577 \\
\midrule
Stack Exchange (Train) & 0.540 & 0.554 & 0.540 & 0.562 & 0.540 & 0.567 & 0.540 & 0.574 \\
Stack Exchange (Val) & 0.518 & 0.523 & 0.518 & 0.533 & 0.518 & 0.534 & 0.518 & 0.538 \\
\bottomrule
\end{tabular}
\end{table}

% Table 8
\begin{table}[!h]
\centering
\caption{\textbf{Calibration-test $p$-values (Diff), full-scale held-out generation} (same subsets as \Cref{tab:posthoc-aucs-small-pile}). Symbols as in \Cref{tab:posthoc-pvalues-subsample}. With the distribution-shift artifact removed, the test aligns far better with ground truth than at benchmark scale (\Cref{tab:posthoc-pvalues-subsample}): 23/32 vs.\ 12/32 correct, with Hacker News and Europarl train splits now flagged on larger models, though Common Crawl and Stack Exchange remain undetected.}
\label{tab:posthoc-p-values-new-mias}
\scriptsize
\resizebox{\linewidth}{!}{%
\begin{tabular}{@{\extracolsep{\fill}}l|cc|cc|cc|cc|c}
\toprule
& \multicolumn{2}{c|}{\textbf{Pythia 410M}} & \multicolumn{2}{c|}{\textbf{Pythia 1.4B}} & \multicolumn{2}{c|}{\textbf{Pythia 2.8B}} & \multicolumn{2}{c|}{\textbf{Pythia 6.9B}} & \textbf{Corr./Tot.} \\
\cmidrule(lr){2-3} \cmidrule(lr){4-5} \cmidrule(lr){6-7} \cmidrule(lr){8-9}
Dataset & Train & Val & Train & Val & Train & Val & Train & Val & \\
\midrule
Pile CC
& $0.775$ \rminus & $0.728$ \gminus
& $0.829$ \rminus & $0.647$ \gminus
& $0.580$ \rminus & $0.717$ \gminus
& $0.582$ \rminus & $0.547$ \gminus
& 4/8 \\
Europarl
& $0.184$ \rminus & $0.882$ \gminus
& $0.000$ \gplus & $0.675$ \gminus
& $0.000$ \gplus & $0.696$ \gminus
& $0.000$ \gplus & $0.350$ \gminus
& 7/8 \\
Hacker News
& $0.002$ \gplus & $0.744$ \gminus
& $0.007$ \gplus & $0.958$ \gminus
& $0.000$ \gplus & $0.774$ \gminus
& $0.002$ \gplus & $0.876$ \gminus
& 8/8 \\
Stack Exch.
& $0.598$ \rminus & $0.830$ \gminus
& $0.254$ \rminus & $0.435$ \gminus
& $0.216$ \rminus & $0.377$ \gminus
& $0.116$ \rminus & $0.247$ \gminus
& 4/8 \\
\bottomrule
\end{tabular}%
}
\end{table}

Calibration test $p$-values in \Cref{tab:posthoc-pvalues-subsample} are correspondingly inconsistent. For Common Crawl (410M), train yields $p=0.583$ while validation yields $p=0.396$, both above the rejection threshold.
Europarl produces $p\approx1.000$ across splits and model sizes. Hacker News (Val) produces extremely small values such as $p=0.007$ for 1.4B despite being a non-member split.

To test it is this failure intrinsic to Post-Hoc DI, or just a symptom of limited data at benchmark scale, we re-run the calibration stage on the authors' released held-out sets, whose generator is trained on nearly the full Pile subsets. Text-only AUCs collapse to around 55\% (\Cref{tab:posthoc-aucs-small-pile}, e.g.\ Common Crawl Base: $0.548$), so the distribution shift that dominated the benchmark-scale setting (\Cref{tab:posthoc-aucs-subsample}) largely vanishes. With the artifact gone, the calibration test recovers, rising from 12/32 to 23/32 correct $p$-values (\Cref{tab:posthoc-p-values-new-mias} vs.\ \Cref{tab:posthoc-pvalues-subsample}). Post-Hoc DI is underpowered at benchmark scale, working only with enough data to train a well-matched synthetic generator.

\begin{figure}[H]
    \centering
    \includegraphics[width=0.45\linewidth]{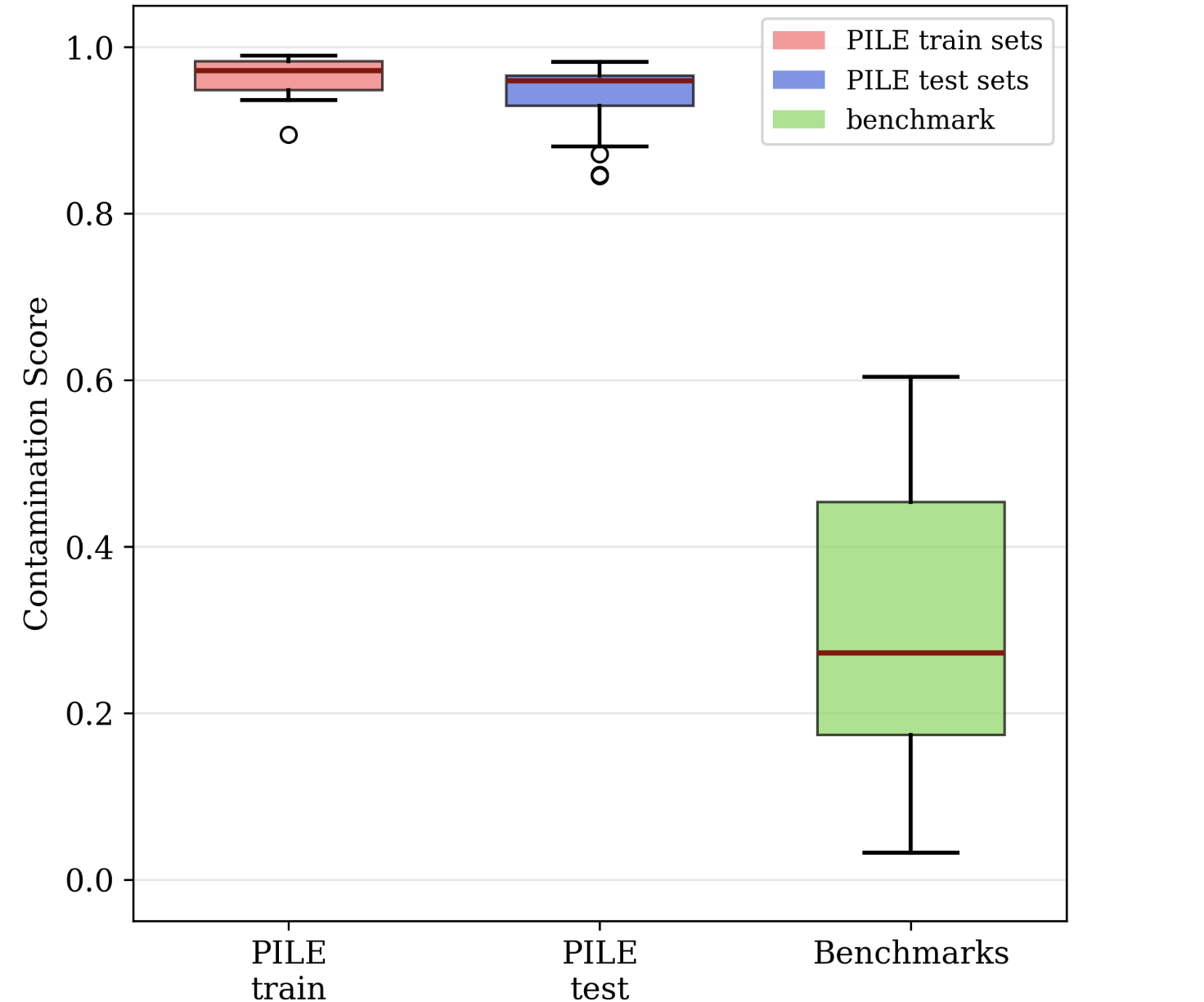}
    \caption{\textbf{CoDeC contamination scores on Pythia.}
    Evaluation-only benchmarks score lower than pre-training corpora (reproducing prior work), but train and test splits of the same corpus score nearly identically.}
    \label{fig:pythia_all_datasets}
\end{figure}

\paragraph{CoDeC.}
CoDeC exhibits a different limitation than the DI-based methods: it preserves broad provenance differences, but lacks the resolution needed for split-level auditing. In \Cref{fig:pythia_all_datasets}, contamination scores for known pre-training corpora are substantially higher than for evaluation-only benchmarks, reproducing prior findings. However, train and test splits within the same dataset yield only marginal differences, indicating difficulty distinguishing near-IID conditions.
\begin{figure}[!h]
    \centering
    \includegraphics[width=1\linewidth]{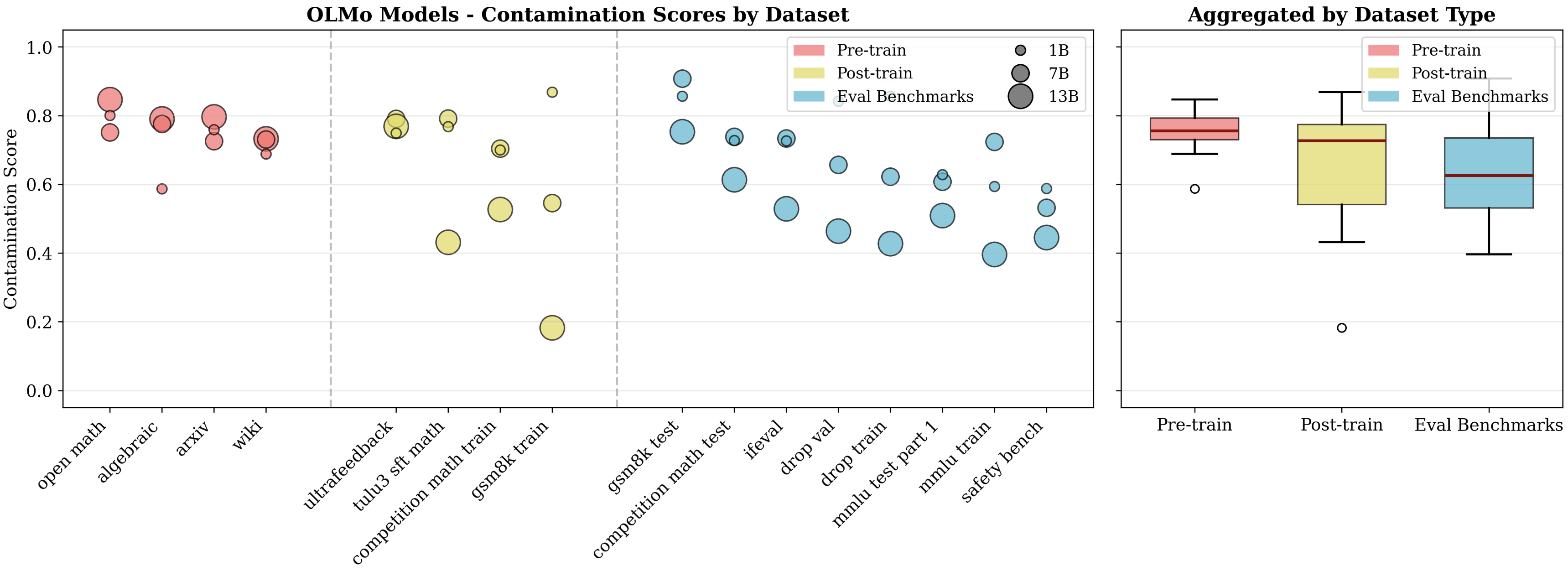}
    \caption{\textbf{CoDeC scores for OLMo~2 (instruction-tuned) grouped by data provenance.} Pre-training sources receive the highest contamination scores, post-training datasets used during instruction tuning are intermediate, and evaluation-only benchmarks score lowest. This coarse ordering is consistent with known data membership, but the overlap between adjacent groups limits the method's utility for certifying individual benchmark integrity.}
    \label{fig:olmo_pretrain_bench_combined}
\end{figure}
In instruction-tuned models (\Cref{fig:olmo_pretrain_bench_combined}), aggregation by dataset provenance reveals a coarse ordering: pre-training sources receive the highest scores, post-training datasets are intermediate, and evaluation-only benchmarks are lowest. At the level of individual datasets and model sizes, however, the scores remain variable and overlapping. CoDeC is therefore best interpreted as a coarse contamination indicator rather than a precise split-level detector. Used comparatively---many models on one dataset, flagging outliers---it still offers a useful relative signal of likely contamination.

\section{Limitation and Future Work}
    
Our study has several limitations that also point to future work. Our results come from configurations run under a shared interface with default hyperparameters rather than exhaustive per-task tuning, which might otherwise improve the effectiveness of the evaluated methods. Our open-weight audit reaches 27B parameters and centres on math, QA, and knowledge benchmarks in English, with additional Polish (PLLuM) and medical models; larger open models, other languages and modalities, and free-form generation benchmarks are not covered, and behaviour may differ at frontier scale. For closed industry models we lack verified membership, so that evidence is comparative across model families rather than definitive. Finally, we characterise \emph{when} and \emph{why} current methods fail in the wild, but do not propose a new detector that overcomes these failures. Quantifying how much distribution shift or how little data each method can tolerate, and designing auditing tools robust to both, remains open.
\section{Conclusions}

In this work, we examined the reliability of contamination detection methods by moving from the controlled settings of prior literature to the realistic, "in-the-wild" regime of modern instruction-tuned models. We stress-tested three leading detection paradigms against the complexities of post-training mixtures and opaque data provenance.
Our findings reveal that the transition from academic validation to practical auditing is fraught with challenges:

\begin{itemize}
    \item \textbf{The I.I.D. assumption is a critical vulnerability.} We found that \textit{LLM Dataset Inference}~\cite{maini2024llmdatasetinferencedid} is highly sensitive to distribution shifts. While effective when a perfect I.I.D. validation set exists, it yields significant false positives when using standard benchmark test splits as validation proxies (as seen with DROP and MMLU). This makes it dangerous to use as a sole arbiter for leaderboard integrity without verifying distributional alignment.
    
    \item \textbf{Benchmark-scale datasets are too small for generative methods.} While \textit{Post-Hoc Dataset Inference}~\cite{zhao2025posthocdi} theoretically removes the need for natural validation data, we showed it is impractical for standard benchmarks. Unlike massive pre-training corpora, benchmarks like GSM8K lack the text volume required to effectively train the required generators.
    
    \item \textbf{Detection granularity is limited.} \textit{CoDeC}~\cite{zawalski2025detectingdatacontaminationllms} provides a valuable coarse-grained signal—effectively separating pre-training sources from held-out tasks in aggregate. However, its high variance at the individual dataset train level (\Cref{fig:olmo_pretrain_bench_combined}) limits its utility for certifying specific benchmark results.
\end{itemize}

We conclude that there is currently no "silver bullet" for detecting contamination in the wild. Consequently, statistical auditing cannot yet fully replace \textbf{transparent data provenance}.

\begin{credits}
\subsubsection{\ackname}

This research was supported by the Polish National Science Centre (NCN) within grants no. 2025/57/N/ST6/04025.
We gratefully acknowledge Polish high-performance computing infrastructure PLGrid, HPC Center: ACK Cyfronet AGH, for providing computer facilities and support within computational grant no. PLG/2024/017781 and PLG/2025/018634.

\subsubsection{\discintname}
The authors have no competing interests to declare that are relevant to the content of this article.
\end{credits}

\bibliographystyle{splncs04}
\bibliography{reference}

\begin{thebibliography}{10}
\providecommand{\url}[1]{\texttt{#1}}
\providecommand{\urlprefix}{URL }
\providecommand{\doi}[1]{https://doi.org/#1}

\bibitem{alonso2024medexpqamultilingualbenchmarkinglarge}
Alonso, I., Oronoz, M., Agerri, R.: Medexpqa: Multilingual benchmarking of large language models for medical question answering. ArXiv  \textbf{abs/2404.05590} (2024)

\bibitem{balloccu2024leakcheatrepeatdata}
Balloccu, S., Schmidtová, P., Lango, M., Dušek, O.: Leak, cheat, repeat: Data contamination and evaluation malpractices in closed-source llms. ArXiv  \textbf{abs/2402.03927} (2024)

\bibitem{biderman2023pythiasuiteanalyzinglarge}
Biderman, S., Schoelkopf, H., Anthony, Q., Bradley, H., O'Brien, K., et~al.: Pythia: A suite for analyzing large language models across training and scaling. ArXiv  \textbf{abs/2304.01373} (2023)

\bibitem{brown2020languagemodelsfewshotlearners}
Brown, T.B., Mann, B., Ryder, N., Subbiah, M., Kaplan, J., et~al.: Language models are few-shot learners. ArXiv  \textbf{abs/2005.14165} (2020)

\bibitem{cobbe2021gsm8k}
Cobbe, K., Kosaraju, V., Bavarian, M., Chen, M., Jun, H., et~al.: Training verifiers to solve math word problems. ArXiv  \textbf{abs/2110.14168} (2021)

\bibitem{cui2023ultrafeedback}
Cui, G., Yuan, L., Ding, N., Yao, G., Zhu, W., et~al.: Ultrafeedback: Boosting language models with high-quality feedback. ArXiv  \textbf{abs/2310.01377} (2023)

\bibitem{dekoninck2024constatperformancebasedcontaminationdetection}
Dekoninck, J., Müller, M.N., Vechev, M.: Constat: Performance-based contamination detection in large language models. ArXiv  \textbf{abs/2405.16281} (2024)

\bibitem{deng2024investigatingdatacontaminationmodern}
Deng, C., Zhao, Y., Tang, X., et~al.: Investigating data contamination in modern benchmarks for large language models. ArXiv  \textbf{abs/2311.09783} (2024)

\bibitem{dua2019drop}
Dua, D., Wang, Y., Dasigi, P., Stanovsky, G., Singh, S., et~al.: {DROP}: A reading comprehension benchmark requiring discrete reasoning over paragraphs. ArXiv  \textbf{abs/1903.00161} (2019)

\bibitem{dubinski2025cdi}
Dubi{\'n}ski, J., Kowalczuk, A., Boenisch, F., Dziedzic, A.: Cdi: Copyrighted data identification in diffusion models. In: Proceedings of the Computer Vision and Pattern Recognition Conference. pp. 18674--18684 (2025)

\bibitem{dubinski2024towards}
Dubi{\'n}ski, J., Kowalczuk, A., Pawlak, S., Rokita, P., Trzci{\'n}ski, T., Morawiecki, P.: Towards more realistic membership inference attacks on large diffusion models. In: Proceedings of the IEEE/CVF Winter Conference on Applications of Computer Vision. pp. 4860--4869 (2024)

\bibitem{gao2020pile800gbdatasetdiverse}
Gao, L., Biderman, S., Black, S., Golding, L., Hoppe, T., et~al.: The pile: An 800gb dataset of diverse text for language modeling. ArXiv  \textbf{abs/2101.00027} (2020)

\bibitem{golchin2024timetravelllmstracing}
Golchin, S., Surdeanu, M.: Time travel in llms: Tracing data contamination in large language models. ArXiv  \textbf{abs/2308.08493} (2024)

\bibitem{hendrycks2021mmlu}
Hendrycks, D., Burns, C., Basart, S., Zou, A., Mazeika, M., et~al.: Measuring massive multitask language understanding. ArXiv  \textbf{abs/2009.03300} (2021)

\bibitem{hendrycks2021math}
Hendrycks, D., Burns, C., Kadavath, S., Arora, A., et~al.: Measuring mathematical problem solving with the {MATH} dataset. ArXiv  \textbf{abs/2103.03874} (2021)

\bibitem{jin2020disease}
Jin, D., Pan, E., Oufattole, N., Weng, W.H., Fang, H., et~al.: What disease does this patient have? a large-scale open domain question answering dataset from medical exams. ArXiv  \textbf{abs/2009.13081} (2020)

\bibitem{jin-etal-2019-pubmedqa}
Jin, Q., Dhingra, B., Liu, Z., Cohen, W., Lu, X.: {P}ub{M}ed{QA}: A dataset for biomedical research question answering. In: Proceedings of the 2019 Conference on Empirical Methods in Natural Language Processing and the 9th International Joint Conference on Natural Language Processing (EMNLP-IJCNLP) (Nov 2019)

\bibitem{kim2024smalllanguagemodelslearn}
Kim, H., Hwang, H., Lee, J., Park, S., Kim, D., et~al.: Small language models learn enhanced reasoning skills from medical textbooks. ArXiv  \textbf{abs/2404.00376} (2024)

\bibitem{kocoń2025pllumfamilypolishlarge}
Kocoń, J., Piasecki, M., Janz, A., Ferdinan, T., Łukasz Radliński, et~al.: Pllum: A family of polish large language models. ArXiv  \textbf{abs/2511.03823} (2025)

\bibitem{kowalczuk2025privacy}
Kowalczuk, A., Dubi{\'n}ski, J., Boenisch, F., Dziedzic, A.: Privacy attacks on image autoregressive models. arXiv preprint arXiv:2502.02514  (2025)

\bibitem{Lambert2024TLU3P}
Lambert, N., Morrison, J.D., Pyatkin, V., Huang, S., et~al.: T{\"u}lu 3: Pushing frontiers in open language model post-training. ArXiv  \textbf{abs/2411.15124} (2024)

\bibitem{lee2022deduplicatingtrainingdatamakes}
Lee, K., Ippolito, D., Nystrom, A., Zhang, C., Eck, D., et~al.: Deduplicating training data makes language models better. ArXiv  \textbf{abs/2107.06499} (2022)

\bibitem{maini2024llmdatasetinferencedid}
Maini, P., Jia, H., Papernot, N., Dziedzic, A.: Llm dataset inference: Did you train on my dataset? ArXiv  \textbf{abs/2406.06443} (2024)

\bibitem{olmo20252olmo2furious}
OLMo, T., Walsh, P., Soldaini, L., Groeneveld, D., Lo, K., et~al.: Olmo 2: Furious. ArXiv  \textbf{abs/2501.00656} (2025)

\bibitem{pal2022medmcqalargescalemultisubject}
Pal, A., Umapathi, L.K., Sankarasubbu, M.: Medmcqa : A large-scale multi-subject multi-choice dataset for medical domain question answering. ArXiv  \textbf{abs/2203.14371} (2022)

\bibitem{sallinen2025llamameditron}
Sallinen, A., Solergibert, A.J., Zhang, M., Boy{\'e}, G.B., et~al.: Llama-3-meditron: An open-weight suite of medical {LLM}s based on llama-3.1. In: Workshop on Large Language Models and Generative AI for Health at AAAI 2025 (2025)

\bibitem{sellergren2025medgemmatechnicalreport}
Sellergren, A., Kazemzadeh, S., Jaroensri, T., Kiraly, A., Traverse, M., et~al.: Medgemma technical report. ArXiv  \textbf{abs/2507.05201} (2025)

\bibitem{shokri2017membership}
Shokri, R., Stronati, M., Song, C., Shmatikov, V.: Membership inference attacks against machine learning models. ArXiv  \textbf{abs/1610.05820} (2017). \doi{10.1109/SP.2017.41}

\bibitem{Neeto-1.0-8b}
Verma, S.: Neeto: A specialized medical llm for neet-pg/ukmle/usmle preparation (2025), \url{https://huggingface.co/S4nfs/Neeto-1.0-8b}

\bibitem{zawalski2025detectingdatacontaminationllms}
Zawalski, M., Boubdir, M., Bałazy, K., Nushi, B., Ribalta, P.: Detecting data contamination in llms via in-context learning. ArXiv  \textbf{abs/2510.27055} (2025)

\bibitem{zhao2025posthocdi}
Zhao, B., Maini, P., Boenisch, F., Dziedzic, A.: Unlocking post-hoc dataset inference with synthetic data. ArXiv  \textbf{abs/2506.15271} (2025)

\end{thebibliography}

\clearpage
\appendix
\section{Detailed Method Overviews}
\label{sec:method-overviews}

This section provides step-by-step procedural summaries for the three detection paradigms evaluated in the main text.

\subsubsection{LLM Dataset Inference}
\label{sec:llm-di-overview}

LLM Dataset Inference~\cite{maini2024llmdatasetinferencedid} detects training-data membership at the \emph{dataset} level by aggregating weak per-sample membership inference signals and applying a statistical test.
The procedure consists of four stages.

\textbf{Stage~0 (Data preparation).}
The auditor requires two datasets drawn from the same distribution: a \emph{suspect set}~$\mathcal{D}_{\text{sus}}$, hypothesised to have been used for training, and a \emph{validation set}~$\mathcal{D}_{\text{val}}$, known to be unseen by the model.
Both sets are randomly partitioned into non-overlapping A~and~B splits.

\textbf{Stage~1 (Feature extraction).}
For every sample in the A-partitions, a battery of membership inference attacks (MIAs) is executed against the suspect LLM~$f_\theta$ to produce a feature vector.
These features span loss-based metrics (perplexity, Min-$k$\%), perturbation-based scores (synonym substitution, character-level edits), reference-model comparisons, and compression-based ratios---52 scores in total.

\textbf{Stage~2 (Correlation learning).}
A linear regressor is trained on the A-split features, assigning label~$0$ to suspect samples and label~$1$ to validation samples.
The learned weights identify which MIA features carry positive membership signal for the given data distribution, since no single MIA works across all distributions.

\textbf{Stage~3 (Statistical test).}
The regressor produces per-sample scores on the held-out B-splits.
A one-sided $t$-test assesses whether the suspect-set scores are systematically lower (more member-like) than the validation-set scores, yielding a $p$-value under the null hypothesis~$H_0$ that the suspect set was not used for training.

\textbf{Critical assumption.}
The validation set must be both IID with the suspect set and genuinely unseen during training.
When this assumption is violated---for example when using a benchmark test split as validation for its train split---the method can produce false positives, as demonstrated in Task~2 of our evaluation.

\subsubsection{Post-Hoc Dataset Inference~}
\label{sec:posthoc-di-overview}

Post-Hoc Dataset Inference~\cite{zhao2025posthocdi}  removes the requirement for a natural IID validation set by synthetically generating one.
The method operates in two phases.

\textbf{Phase~1: Held-out data generation.}
The suspect documents are segmented into short text snippets, shuffled, and randomly split into a \emph{generator training set} and a \emph{generator inference set}.
A small causal language model is fine-tuned (via LoRA) on the training set using a next-token prediction objective.
Each snippet in the inference set is then split into a prefix and a suffix.
The trained generator produces synthetic suffixes conditioned on the prefixes.
The original (real) suffixes constitute the suspect set~$\mathcal{D}_{\text{sus}}$; the synthetic suffixes constitute the held-out set~$\mathcal{D}_{\text{val}}$.
Because every real--synthetic pair shares a common prefix, the two sets are positionally aligned, minimising distributional shift.

\textbf{Phase~2: Post-hoc calibration.}
Both sets are split into train and test subsets.
Two classifiers are trained on the training subsets:
(i)~a \emph{text-only classifier}~$c_{\text{text}}(x)$ that distinguishes real from synthetic text based on textual features alone, and
(ii)~a \emph{combined classifier}~$c_{\text{comb}}(x, \mathrm{MIA}(f(x)))$ that additionally incorporates MIA-derived features from the suspect LLM~$f$.
The key insight is that any performance advantage of $c_{\text{comb}}$ over $c_{\text{text}}$ must stem from genuine membership signals rather than the real-vs-synthetic distributional shift.
A one-sided $t$-test on the test subsets checks whether this advantage is statistically significant; rejection of~$H_0$ indicates that the suspect set was used for training.

\textbf{Critical assumption.}
The generator must be trained on sufficient data to produce realistic synthetic text.
When trained on only ${\sim}2{,}000$ benchmark-sized samples, the generator is too weak: the text-only classifier achieves high AUC, the real-vs-synthetic gap dominates, and the calibration test becomes uninformative.

\subsubsection{CoDeC: Contamination Detection via Context}
\label{sec:codec-overview}

CoDeC~\cite{zawalski2025detectingdatacontaminationllms} detects dataset-level contamination by measuring how in-context examples from the same dataset affect the model's prediction confidence.
The procedure operates on each sample independently before aggregating.

\textbf{Step~1 (Baseline prediction).}
For each sample $x \in \mathcal{D}$, compute the model's average per-token log-likelihood on~$x$ without any preceding context.

\textbf{Step~2 (In-context prediction).}
Sample $n$ additional examples from $\mathcal{D} \setminus \{x\}$, prepend them to~$x$ as in-context demonstrations, and recompute the model's average log-likelihood on~$x$ in this extended context.

\textbf{Step~3 (Score computation).}
Compute the per-sample confidence difference $\Delta(x) = \text{logprob}_{\text{ICL}}(x) - \text{logprob}_{\text{baseline}}(x)$.
For unseen data, in-context examples typically improve confidence ($\Delta > 0$), since they provide useful distributional cues such as format, vocabulary, and style.
For memorised data, context can disrupt memorisation patterns, \emph{reducing} confidence ($\Delta < 0$).

\textbf{Step~4 (Aggregation).}
The dataset contamination score is
\[
  S(\mathcal{D}) \;=\; \frac{1}{N}\sum_{i=1}^{N} \mathbf{1}\!\bigl[\Delta(x_i) < 0\bigr],
\]
the fraction of samples for which context reduces confidence.
Scores above $80\%$ indicate strong contamination evidence; scores below $60\%$ suggest no contamination; intermediate values require cross-model comparison for reliable interpretation.

\textbf{Key properties.}
CoDeC requires no external reference data, held-out sets, or dataset-specific calibration.
It produces percentage-based scores that are directly interpretable and model-agnostic, needing only gray-box access to token log-probabilities.
The method is also computationally efficient, requiring only two forward passes per sample.

\section{Numerical values for evaluation experiments}
We present numerical values which stand for symbols in tables in main text.

    \begin{table}[H]
    \caption{Numeric contamination-detection scores on train and test splits of the Pile (Limited Data). LLM DI and Post-Hoc DI entries are p-values; CoDeC entries are contamination scores.}
    \label{tab:task1_results_numeric}
    \centering
    \resizebox{\textwidth}{!}{%
    \begin{tabular}{lcccc}
    \toprule
    & \multicolumn{2}{c}{Pile CC} & \multicolumn{2}{c}{Pile Europarl} \\
    \cmidrule(lr){2-3} \cmidrule(lr){4-5}
    & Train & Test & Train & Test \\
    \textit{Pythia sizes} & \multicolumn{1}{c}{0.4B/1.4B/2.8B/6.9B} & \multicolumn{1}{c}{0.4B/1.4B/2.8B/6.9B} & \multicolumn{1}{c}{0.4B/1.4B/2.8B/6.9B} & \multicolumn{1}{c}{0.4B/1.4B/2.8B/6.9B} \\
    \midrule
    LLM DI & 1.00 / 1.00 / 0.95 / 1.00 & 1.00 / 1.00 / 1.00 / 1.00 & 0.00 / 0.00 / 0.00 / 0.00 & 1.00 / 1.00 / 1.00 / 1.00 \\
    Post-Hoc DI & 0.58 / 0.44 / 0.40 / 0.39 & 0.40 / 0.28 / 0.26 / 0.24 & 1.00 / 1.00 / 1.00 / 1.00 & 1.00 / 1.00 / 1.00 / 1.00 \\
    CoDeC & 0.88 / 0.86 / 0.87 / 0.86 & 0.84 / 0.84 / 0.84 / 0.86 & 0.92 / 0.94 / 0.96 / 0.97 & 0.96 / 0.97 / 0.97 / 0.98 \\
    \midrule
    \midrule
    & \multicolumn{2}{c}{Pile Hacker News} & \multicolumn{2}{c}{Pile Stack Exchange} \\
    \cmidrule(lr){2-3} \cmidrule(lr){4-5}
    & Train & Test & Train & Test \\
    \textit{Pythia sizes} & \multicolumn{1}{c}{0.4B/1.4B/2.8B/6.9B} & \multicolumn{1}{c}{0.4B/1.4B/2.8B/6.9B} & \multicolumn{1}{c}{0.4B/1.4B/2.8B/6.9B} & \multicolumn{1}{c}{0.4B/1.4B/2.8B/6.9B} \\
    \midrule
    LLM DI & 0.01 / 0.01 / 0.02 / 0.00 & 1.00 / 1.00 / 0.98 / 1.00 & 0.00 / 0.00 / 0.01 / 0.00 & 1.00 / 1.00 / 1.00 / 1.00 \\
    Post-Hoc DI & 0.45 / 0.44 / 0.32 / 0.39 & 0.05 / 0.01 / 0.01 / 0.01 & 0.71 / 0.71 / 0.70 / 0.75 & 0.82 / 0.81 / 0.75 / 0.79 \\
    CoDeC & 0.90 / 0.95 / 0.97 / 0.97 & 0.91 / 0.94 / 0.95 / 0.96 & 0.94 / 0.96 / 0.95 / 0.95 & 0.88 / 0.87 / 0.85 / 0.84 \\
    \bottomrule
    \end{tabular}%
    }
    \end{table}

\begin{table}[H]
    \caption{Numeric contamination-detection scores for OLMo~2. LLM DI and Post-Hoc DI entries are p-values; CoDeC entries are contamination scores.}
    \label{tab:task2_results_numeric}
    \centering
    \resizebox{1\textwidth}{!}{%
    \begin{tabular}{lcccccc}
    \toprule
    & \multicolumn{2}{c}{GSM8K} & \multicolumn{2}{c}{MMLU} & \multicolumn{2}{c}{DROP} \\
    \cmidrule(lr){2-3} \cmidrule(lr){4-5} \cmidrule(lr){6-7}
    & Train & Test & Train & Test & Train & Test \\
    \textit{OLMo 2 version} & \multicolumn{2}{c}{1B / 7B / 13B} & \multicolumn{2}{c}{1B / 7B / 13B} & \multicolumn{2}{c}{1B / 7B / 13B} \\
    \midrule
    LLM DI & 0.00 / 0.00 / 0.00 & 1.00 / 1.00 / 1.00 & 0.00 / 0.00 / 0.00 & 1.00 / 0.97 / 0.99 & 0.00 / 0.00 / 0.00 & 1.00 / 1.00 / 0.51 \\
    Post-Hoc DI & 0.80 / 0.54 / 0.06 & n/a / n/a / n/a & 0.68 / 0.75 / 0.61 & 0.74 / 0.90 / 0.76 & 1.00 / 1.00 / 1.00 & 0.99 / 0.99 / 0.99 \\
    CoDeC & 0.86 / 0.56 / 0.17 & 0.87 / 0.92 / 0.76 & 0.59 / 0.73 / 0.40 & 0.60 / 0.58 / 0.52 & 0.87 / 0.59 / 0.37 & 0.81 / 0.67 / 0.45 \\
    \bottomrule
    \end{tabular}%
    }
    \end{table}

    \begin{table}[H]
    \centering
    \caption{Contamination detection scores for medical LLMs (numeric values).}
    \label{tab:medical-scores-numeric}
    \footnotesize
    \begin{tabular}{@{\extracolsep{\fill}}ll|cc|cc}
    \toprule
    & & \multicolumn{2}{c|}{\textbf{MedQA-USMLE}} & \multicolumn{2}{c}{\textbf{MedMCQA}} \\
    \cmidrule(lr){3-4} \cmidrule(lr){5-6}
    & & Train & Test & Train & Test \\
    \midrule
    \multirow{2}{*}{\makecell{medgemma\\4B / 27B}} & LLM DI & 0.17 / 0.25 & 1.00 / 0.73 & 0.17 / 0.01 & 1.00 / 0.13 \\
     & CoDeC & 0.00 / 0.00 & 0.00 / 0.00 & 0.00 / 0.00 & 0.00 / 0.00 \\
    \midrule
    \multirow{2}{*}{\makecell{Meditron3\\8B / 9B}} & LLM DI & 0.69 / 0.98 & 0.63 / 0.83 & 0.00 / 0.05 & 1.00 / 1.00 \\
     & CoDeC & 0.66 / 0.29 & 0.64 / 0.31 & 0.37 / 0.32 & 0.34 / 0.28 \\
    \midrule
    \multirow{2}{*}{\makecell{meerkat\\7B / 8B}} & LLM DI & 0.48 / 0.99 & 1.00 / 0.97 & 0.03 / 0.26 & 1.00 / 1.00 \\
     & CoDeC & 0.71 / 0.65 & 0.71 / 0.63 & 0.47 / 0.45 & 0.46 / 0.48 \\
    \midrule
    \multirow{2}{*}{\makecell{Neeto-1.0\\8B}} & LLM DI & 0.83 & 0.88 & 0.73 & 1.00 \\
     & CoDeC & 0.52 & 0.51 & 0.43 & 0.48 \\
    \bottomrule
    \end{tabular}
    \end{table}

    \begin{table}[H]
    \centering
    \caption{Contamination detection scores for PLLuM models on SFT data (numeric values). base* = updated base checkpoint (250801).}
    \label{tab:pllum-scores-sft-numeric}
    \footnotesize
    \begin{tabular}{@{\extracolsep{\fill}}ll|cc}
    \toprule
    \multicolumn{4}{c}{\textbf{Automatic SFT}} \\
    \midrule
    & & Train & Val \\
    \midrule
    \multirow{3}{*}{\makecell{PLLuM-12B\\base}} & \textit{base / base* / nc-base} & & \\
     & LLM DI & 0.92 / 0.62 / 0.85 & 0.71 / 0.26 / 0.90 \\
     & CoDeC & 0.85 / 0.50 / 0.38 & 0.82 / 0.49 / 0.37 \\
    \midrule
    \multirow{6}{*}{\makecell{PLLuM-12B\\SFT}} & \textit{chat / inst.} & & \\
     & LLM DI & 0.00 / 0.32 & 0.50 / 0.74 \\
     & CoDeC & 0.90 / 0.89 & 0.86 / 0.89 \\
     & \textit{nc-chat / nc-inst.} & & \\
     & LLM DI & 0.46 / 0.81 & 0.18 / 0.16 \\
     & CoDeC & 0.81 / 0.61 & 0.79 / 0.59 \\
    \midrule
    \multirow{3}{*}{\makecell{Llama-PLLuM\\8B base}} & \textit{base / base*} & & \\
     & LLM DI & 0.64 / 0.15 & 0.72 / 0.42 \\
     & CoDeC & 0.95 / 0.55 & 0.96 / 0.57 \\
    \midrule
    \multirow{3}{*}{\makecell{Llama-PLLuM\\8B SFT}} & \textit{chat / inst.} & & \\
     & LLM DI & 0.23 / 0.96 & 0.48 / 0.74 \\
     & CoDeC & 0.93 / 0.94 & 0.92 / 0.93 \\
    \midrule
    \midrule
    \multicolumn{4}{c}{\textbf{Manual SFT}} \\
    \midrule
    & & Train & Val \\
    \midrule
    \multirow{3}{*}{\makecell{PLLuM-12B\\base}} & \textit{base / base* / nc-base} & & \\
     & LLM DI & 0.30 / 0.34 / 0.80 & 0.62 / 0.22 / 0.31 \\
     & CoDeC & 0.65 / 0.54 / 0.48 & 0.69 / 0.52 / 0.50 \\
    \midrule
    \multirow{6}{*}{\makecell{PLLuM-12B\\SFT}} & \textit{chat / inst.} & & \\
     & LLM DI & 0.91 / 0.82 & 0.39 / 0.29 \\
     & CoDeC & 0.78 / 0.66 & 0.79 / 0.77 \\
     & \textit{nc-chat / nc-inst.} & & \\
     & LLM DI & 0.91 / 0.51 & 0.27 / 0.60 \\
     & CoDeC & 0.80 / 0.59 & 0.81 / 0.64 \\
    \midrule
    \multirow{3}{*}{\makecell{Llama-PLLuM\\8B base}} & \textit{base / base*} & & \\
     & LLM DI & 0.82 / 0.65 & 0.64 / 0.26 \\
     & CoDeC & 0.75 / 0.60 & 0.83 / 0.57 \\
    \midrule
    \multirow{3}{*}{\makecell{Llama-PLLuM\\8B SFT}} & \textit{chat / inst.} & & \\
     & LLM DI & 0.91 / 0.80 & 0.65 / 0.23 \\
     & CoDeC & 0.84 / 0.74 & 0.90 / 0.81 \\
    \bottomrule
    \end{tabular}
    \end{table}

    \begin{table}[H]
    \centering
    \caption{Contamination detection scores for PLLuM models on alignment data (numeric values). base* = updated base checkpoint (250801).}
    \label{tab:pllum-scores-alignment-numeric}
    \footnotesize
    \begin{tabular}{@{\extracolsep{\fill}}ll|cc}
    \toprule
    \multicolumn{4}{c}{\textbf{Alignment}} \\
    \midrule
    & & Train & Test \\
    \midrule
    \multirow{3}{*}{\makecell{PLLuM-12B\\base}} & \textit{base / base* / nc-base} & & \\
     & LLM DI & 0.00 / 0.00 / 0.00 & 0.17 / 0.67 / 0.96 \\
     & CoDeC & 0.28 / 0.39 / 0.25 & 0.43 / 0.45 / 0.29 \\
    \midrule
    \multirow{6}{*}{\makecell{PLLuM-12B\\SFT}} & \textit{chat / inst.} & & \\
     & LLM DI & 0.00 / 0.00 & 0.85 / 0.21 \\
     & CoDeC & 0.80 / 0.39 & 0.79 / 0.56 \\
     & \textit{nc-chat / nc-inst.} & & \\
     & LLM DI & 0.00 / 0.00 & 0.94 / 0.81 \\
     & CoDeC & 0.85 / 0.42 & 0.84 / 0.51 \\
    \midrule
    \multirow{3}{*}{\makecell{Llama-PLLuM\\8B base}} & \textit{base / base*} & & \\
     & LLM DI & 0.00 / 0.00 & 0.12 / 0.68 \\
     & CoDeC & 0.37 / 0.36 & 0.47 / 0.47 \\
    \midrule
    \multirow{3}{*}{\makecell{Llama-PLLuM\\8B SFT}} & \textit{chat / inst.} & & \\
     & LLM DI & 0.00 / 0.00 & 0.97 / 0.84 \\
     & CoDeC & 0.83 / 0.49 & 0.85 / 0.64 \\
    \bottomrule
    \end{tabular}
    \end{table}

\section{Time analysis}

    \begin{table}[H]
    \centering
    \caption{\textbf{Average run duration for detection methods.} Standard deviation is reported across different model sizes (Pythia, OLMo 2) and datasets (Pile subsets, GSM8K, DROP and MMLU). All values are normalized to seconds for comparability. The runs were executed on Nvidia A100 Nvidia GH200.}
    \label{tab:run-duration}
    \footnotesize
    \begin{tabular*}{\linewidth}{@{\extracolsep{\fill}}lr}
    \toprule
    \textbf{Method} & \textbf{Average Duration [s]} \\
    \midrule
    Post-Hoc DI & $4566.6 \pm 846.4$ \\
    CoDeC       & $206.4 \pm 232.6$ \\
    LLM DI      & $748.1 \pm 447.4$ \\
    \bottomrule
    \end{tabular*}
    \end{table}

\end{document}